\documentclass{article}
\usepackage{ijcai16}

\usepackage{times}

\usepackage{tipa}
\usepackage{times}
\usepackage{epsfig}
\usepackage{graphicx}
\usepackage{amsmath}
\usepackage{amssymb}
\usepackage{mathrsfs}

\usepackage{balance}
\usepackage{booktabs}
\usepackage{comment}
\usepackage{indentfirst}

\usepackage{lineno}
\usepackage{color}
\usepackage{booktabs}
\usepackage{bbm}
\usepackage{algorithm}
\usepackage{algorithmic}
\usepackage{bm}

\usepackage[pagebackref=true,breaklinks=true,letterpaper=true,colorlinks,citecolor = black,bookmarks=false]{hyperref}

\title{Visual Tracking via Reliable Memories}
\author{$^1$Shu Wang, $^2$Shaoting Zhang, $^3$Wei Liu and $^1$Dimitris N. Metaxas\vspace{1.5mm}\\
\normalsize{$^1$CBIM, Rutgers, The State University of New Jersey, Piscataway, NJ, USA}\\
\normalsize{$^2$Department of Computer Science, UNC Charlotte, Charlotte, NC, USA}\\
\normalsize{$^3$Didi Research, Beijing, China}\\
{\tt\small $^1$\{sw498, dnm\}@cs.rutgers.edu, $^2$szhang16@uncc.edu, $^3$wliu@ee.columbia.edu}
}

\begin{document}

\maketitle

\begin{abstract}\vspace{-2mm}
    In this paper, we propose a novel visual tracking framework that intelligently discovers reliable patterns from a wide range of video to resist drift error
    for long-term tracking tasks.
    First, we design a Discrete Fourier Transform (DFT) based tracker which is able to exploit a large number of tracked samples while still ensures real-time performance.
    Second, we propose a clustering method with temporal constraints to explore and memorize consistent patterns from previous frames, named as ``reliable memories". By virtue of this method, our tracker can utilize uncontaminated information to alleviate drifting issues.
    Experimental results show that our tracker performs favorably against other state-of-the-art methods on benchmark datasets.
    Furthermore, it is significantly competent in handling drifts and able to robustly track challenging long videos over 4000 frames, while most of others lose track at early frames.
\end{abstract}\vspace{-5mm}

\begin{figure*}[htbp]\vspace{-5mm}
\begin{center}\vspace{-5mm}
   \includegraphics[width=0.87\linewidth]{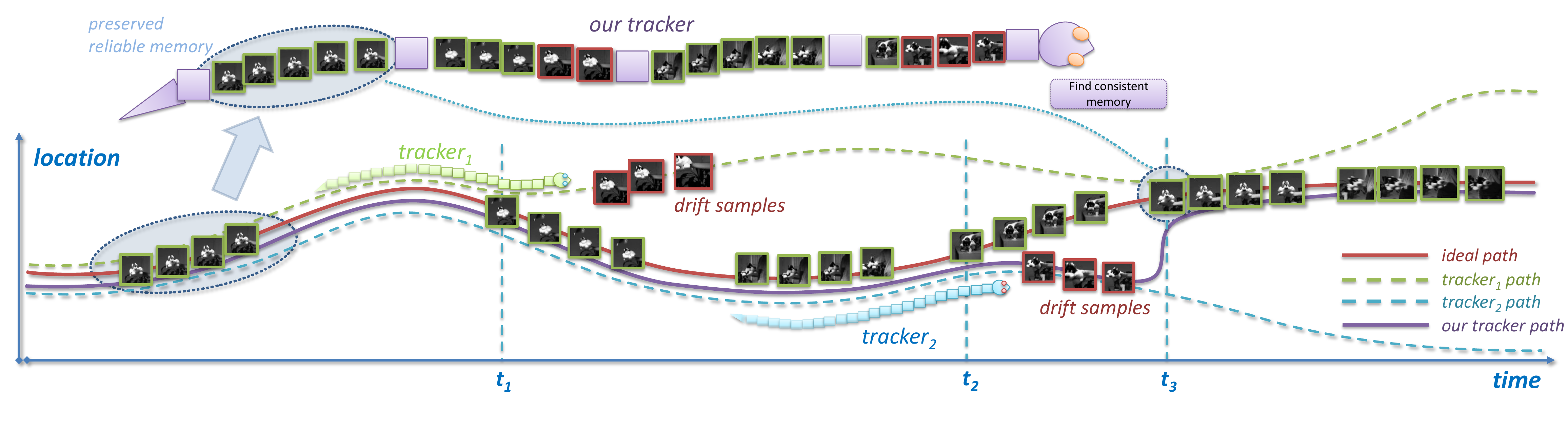}\vspace{-4mm}
\end{center}
\vspace{-1mm}
\vspace{-1mm}
   \caption{
   \footnotesize{
This figure illustrates the basic philosophy of our method. Here we use Snake (video game) as an analogy for learning-rate based visual trackers ($\text{tracker}_1$ and $\text{tracker}_2$): In order to track the target on ideal path, they continuously take in new samples, and forget old ones due to limited memory length. Contrarily, our tracker discovers and preserves multiple temporally constrained clusters as memories, covering a much wider range on the whole sequence.
As shown above, $\text{tracker}_1$, $\text{tracker}_2$ and our tracker depart from the ideal path at time $t_1$ and $t_2$ for drastic target appearance changes. After that, all three trackers absorb a certain amount of drifted samples. With only limited length of memory, $\text{tracker}_2$ can hardly recover from drift error even if familiar target appearance shows up at $t_3$. Similarly, $\text{tracker}_1$ deviates from the ideal path for long and is degraded by drifted samples from time $t_1$ to $t_3$. Even it happens to be close to the ideal path at $t_3$ by chance, without keeping memory on similar samples long before, it still drifts from the ideal path with a high probability.
On the contrary, when similar target appearance occurs at $t_3$, our tracker corrects tracking result with consistent and reliable memories, and recovers from drift error.}
   } \label{fig:idealpath}\vspace{-5mm}
\end{figure*}

\vspace{-0mm}
\section{Introduction}
\vspace{-1mm}
Visual tracking is one of the fundamental and challenging problems in computer vision and artificial intelligence.
Though much progress has been achieved in recent
years~\cite{YilmazJS06,WuLimYang13},
there are still unsolved issues due to its complexity on various factors, such as
illumination and angle changes, clutter background, shape deformation and occlusion.
Extensive studies on visual tracking employ a tracking-by-detection framework and
achieve promising results by extending existing machine learning methods (usually discriminative)
with online learning manner~\cite{avidan2004support,avidan2007ensemble,grabner2008semi,saffari2009line}.
To adaptively model various appearance changes, they deal with a
large amount of samples\footnote{
Here ``samples'' refers to positive (and negative) target patches for trackers based on generative (or discriminative) models.
} at both detection and updating stages.
However, all of them face the same dilemma:
While more samples grant better accuracy and adaptiveness, they also come with higher computational cost and risk of drifting.
In addition to discriminative methods, \cite{ross-ijcv08,mei2011robust,wang2014visual}
utilize generative models with a fixed learning-rate to account for target appearance changes.
The learning-rate is essentially a trade-off between adaptiveness and stability.
However, even with very small rate, former samples' influence on their models still drops exponentially through frames, and drift error may still accumulate.
In order to alleviate drift error, \cite{BabenkoYB11,hare2011struck,zhang2014robust} are designed to exploit hidden structured information around the target region.
Other methods~\cite{Collins-ICCV03,avidan2007ensemble,kwon2010visual} try to avoid drifting
by making the current model a combination of the labeled samples in the first frame and the learned samples from the tracking process.
However, limited number of samples (\emph{e.g.}, the first frame) can be regarded as ``very confident",
which in turn restrict their robustness in long-term challenging tasks.
Recently, several methods~\cite{Bolme-cvpr10,Danelljan-cvpr14,henriques2015tpami} employ Discrete Fourier Transform (DFT)
to perform extremely fast detection and achieve high accuracy with the least computational cost.
However, same as other generative methods, the memory length of their models is
limited by a fixed forgetting rate, and therefore they still suffer from accumulated drift error in long-term tasks.

A very important observation is that, when the tracked target moves smoothly, \emph{e.g.}, without severe occlusion or out-of-plane rotations,
its appearances across frames share high similarity in the feature space (\emph{e.g.}, edge features).
Contrarily, when it undergoes drastic movements such as in/out-of-plane rotations or occlusions, its appearances may not be that similar to previous ones.
Therefore, if we impose a temporal constraint on clustering these samples, such that only temporally adjacent
ones can be grouped together, 
the big clusters with large intra-cluster correlation can indicate the periods when the target experiences small appearance changes.
%
We take human memory as an analogy for these clusters, using \emph{reliable memories} to represent large clusters that have been consistently perceived for a long time.
In this context, earlier memories supported by more samples have higher probability to be reliable than more recent ones with less support, especially when drift error accumulates across frames.
Thus, a tracker may recover from drift error with preference to choose candidates that share high correlation to earlier memories.

Based on these motivations, we propose a novel tracking framework, which efficiently explores self-correlated
appearance clusters across frames, and then preserves reliable memories for long-term robust visual tracking.
First, we design a DFT-based visual tracker, which is capable of retrieving good memories
from a vast number of tracked samples for accurate detection, while still ensures a fast speed for real-time performance.
Second, we propose a novel clustering method with temporal constraints
to discover distinct and reliable memories from previous frames to help our tracker resist drift error.
This method harvests the inherent correlation of the streaming data,
and is guaranteed to converge at a fast speed\footnote{Its computational complexity is $O(n\log n)$, which costs less than 30 ms for $n = 1000$ frames.} by carefully designing upon Integral Image.
To the best of our knowledge, our temporally constrained clustering method is novel
to vision streaming data analysis, and its high converging speed and promising performance
show great potential in online streaming problems. 
Particularly, it is very competent
in discovering clusters (\emph{i.e.}, reliable memories) consisted of uncontaminated sequential
samples that are tracked before,
and grants our tracker remarkable ability to resist drift error.
Experimental results show that our tracker is considerably competent in handling drift error and performs favorably against other state-of-the-art methods on benchmark datasets.
Further, it can robustly track challenging long videos with over $4000$ frames, while most of the others lose track at early frames. 

\vspace{-3mm}
\section{Circulant Structure based Visual Tracking} \label{sec:sec2}
\vspace{-1mm}
Recent works~\cite{Bolme-cvpr10,Henriques-eccv12,Danelljan-cvpr14,henriques2015tpami} achieve
the state-of-the-art tracking accuracy with the least computational
cost
by exploiting the inherent relationship between DFT
and the circulant structure of dense sampling on the target region.
In this section, we briefly introduce these methods that are highly related to our work.

Suppose $\boldsymbol x \in {\mathbb R^{L}} $ is a vector of an image patch with size $M \times N$,
centered at the target ($L = M \times N$), and ${\boldsymbol x_{l}}$ denotes a 2D circular shift
from $\boldsymbol x$ by $m \times n$ ($l$ is an index for all $M \times N$ possible shifts, $ 1 \le l \le L )$.
$\boldsymbol y \in {\mathbb R^{L}}$ is a vector of a designed response map of size $M \times N$ with a
Gaussian pulse centered at the target, too.
$\kappa (\boldsymbol x,\boldsymbol x') =  < \varphi (\boldsymbol x),\varphi (\boldsymbol x') > $ is a positive definite kernel function
defined by mapping $\varphi (\boldsymbol x):{\mathbb R^{L}} \to {\mathbb R^D}$.
We aim to find a linear classifier $f({\boldsymbol x_l}) = {\boldsymbol \omega ^T}\varphi ({{\boldsymbol x}_l}) + b$
that minimizes the Regularized Least Square (RLS) cost function:
\vspace{-1mm}
\begin{equation}\label{eq:rls}
\min \varepsilon (\boldsymbol \omega ) = \mathop {\min }\limits_{\boldsymbol \omega}  \sum\limits_l {{{|| {\boldsymbol{y}_l - f({\boldsymbol{x}_l})} ||}^2}}  + \lambda || f ||_\kappa ^2.\vspace{-2mm}
\end{equation}
The first term is an empirical risk to minimize the difference between
the designed gaussian response $\boldsymbol{y}$ and the mapping ${\boldsymbol x} \rightarrow f_{L}({\boldsymbol x}) \in {\mathbb R^{L}}$, where $f_{l}({\boldsymbol x}) = f({\boldsymbol{x}_{l}})$.
The second term $|| f ||_\kappa$ is a regularization term. It is denoted by $|| f ||_\kappa$ since it
lies in the Kernel Hilbert Space reproduced by $\kappa$.


By Representer Theorem~\cite{ScholkopfHS01}, cost $\varepsilon(\boldsymbol \omega )$ can be minimized by a linear
combination of inputs: $\hat{\boldsymbol \omega} = \sum\limits_{l} {{\alpha _{l}}\varphi ({\boldsymbol{x}_{l}})} $.
By defining kernel matrix $\mathbf K \in {\mathbb R^{ L \times  L}},{\mathbf K(l,l')} = \kappa ({\boldsymbol{x}_{l}},{\boldsymbol{x}_{l'}})$,
a much simpler form for Eq.~\ref{eq:rls} can be derived as:
\vspace{-2mm}
\begin{equation}\label{eq:simplest_rls}
{\min}F(\boldsymbol \alpha ) = \mathop {\min }\limits_{\boldsymbol \alpha}  {(\boldsymbol{y} - \mathbf K\boldsymbol\alpha )^{\rm T}}(\boldsymbol{y} - \mathbf K \boldsymbol\alpha ) + \lambda {\boldsymbol\alpha ^{\rm T}}\mathbf K\boldsymbol\alpha.\vspace{-2mm}
\end{equation}
This function is convex and differentiable, and has a closed form minimizer $\hat{\boldsymbol\alpha} = (\mathbf K + \lambda \mathbf I)^{-1}\boldsymbol{y}$.
As proved in~\cite{Henriques-eccv12}, if the kernel $\kappa$ is unitarily invariant, its kernel matrix $\mathbf K$ is a circulant matrix, that $\mathbf K = C(\boldsymbol{k})$, where vector ${\boldsymbol k} \in R^L$, $k_i = \kappa({\boldsymbol x}, P^i{\boldsymbol x})$.
$P^i$ is a permutation matrix that shifts vectors by $i$-th element(s), $C(\boldsymbol{k})$ is a circulant matrix
from $\boldsymbol{k}$ by concatenating all $L$ possible cyclic shifts of $\boldsymbol{k}$.
and $\hat{\boldsymbol\alpha}$ can be obtained without inverting $(\mathbf K + \lambda \mathbf I)$ by:
\vspace{-2mm}
\begin{equation}\label{eq:alpha_f}
  {\hat{\boldsymbol\alpha}} = {\mathscr F^{ - 1}}(\frac{{\mathscr F({\boldsymbol y})}}{{\mathscr F(\boldsymbol{k}) + \lambda\mathbbm{1} }}),
\vspace{-2mm}
\end{equation}
where $\mathscr F$ and $\mathscr F^{-1}$ are DFT and its inverse, and $\mathbbm{1}$ is
an $n$ by $1$ vector with all entries to be $1$.
Division in Eq.~\ref{eq:alpha_f} is in Fourier domain, and is thus performed element-wise.
In practice, there is no need to compute $\hat {\boldsymbol \alpha}$ from $\hat {\mathbf A}$, since fast detection
can be performed on given image patch $\boldsymbol z$ by $\hat{\boldsymbol{y}} = \mathscr F^{-1}(\mathscr F({\tilde{\boldsymbol{k}}}) \odot\mathscr F({\hat{\boldsymbol\alpha}}))$,
where ${\tilde{\boldsymbol{k}}} \in R^L$ with ${\tilde{k}_l} = \kappa(\boldsymbol{z}, \hat{\boldsymbol{x}}_l)$.
$\hat {\boldsymbol x}$ is the learned target appearance.
Pulse peak in $\hat{\boldsymbol{y}}$ shows the target translation in input image $\boldsymbol{z}$.
Detailed derivation is in~\cite{gray2005toeplitz,rifkin2003regularized,Henriques-eccv12}.

Though recent methods MOSSE~\cite{Bolme-cvpr10}, CSK~\cite{Henriques-eccv12} and ACT~\cite{Danelljan-cvpr14}, have different configurations of kernel functions and features (\emph{e.g.}, dot product kernel $\kappa$ leads to MOSSE, and RBF kernel leads to the latter two),
all of them employ a simple linear combination to learn target appearance model $\{\hat{\boldsymbol x}^p, {\hat{\mathbf A}}^p\}$
at current frame $p$ by
\vspace{-1.0mm}
\begin{equation}\label{eq:x_upd}
    \begin{array}{*{1}{l}}
     {{\hat {\mathbf Q}}^p} = (1 - \gamma ){{\hat {\mathbf Q}}^{p - 1}} + \gamma {\mathbf{ Q}^p}, \;\; {\mathbf Q} = \{{\boldsymbol x}, {\mathbf A}, {\mathbf A}_{N, D}\}.
    \end{array}\vspace{-1mm}
\end{equation}
While CSK updates its classifier coefficients ${\hat {\mathbf A}}^p$ by Eq.~\ref{eq:x_upd} directly, MOSSE and ACT update the numerator ${\hat {\mathbf A}}_N^p$ and
denominator ${\hat {\mathbf A}}_D^p$ of coefficients ${\hat {\mathbf A}}^p$ separately for stability purpose.
The learning-rate $\gamma$ is a trade-off parameter between long memory and model adaptiveness.
After expanding Eq.~\ref{eq:x_upd} we obtain:\vspace{-2mm}
\begin{equation}\label{eq:csk_mosse_upd}\vspace{-1.7mm}
   {{\hat {\mathbf Q}}^p} = \sum\limits_{j = 1}^p {\gamma {{(1 - \gamma )}^{p - j}}{{\mathbf Q}^j}}, \;\;{\mathbf Q} = \{{\boldsymbol x}, {\mathbf A}, {\mathbf A}_{N, D}\}.
\end{equation}
This shows that, all three methods have an exponentially decreasing pattern of memory:
Though the learning-rate $\gamma$ is usually small, \emph{e.g.}, $\gamma = 0.1$, the impact of a sample $\{ {\boldsymbol x}^j, {\mathbf A}^j \}$ at a certain frame $j$ is negligible after $100$ frames ($\gamma (1 - \gamma)^{100} \le 10^{-8}$).
In other words, these learning-rate based trackers are unable to recourse to samples accurately tracked long before
to help resist accumulated drift error.

\begin{figure*}[htbp]
\begin{center}
   \includegraphics[width=0.87\linewidth]{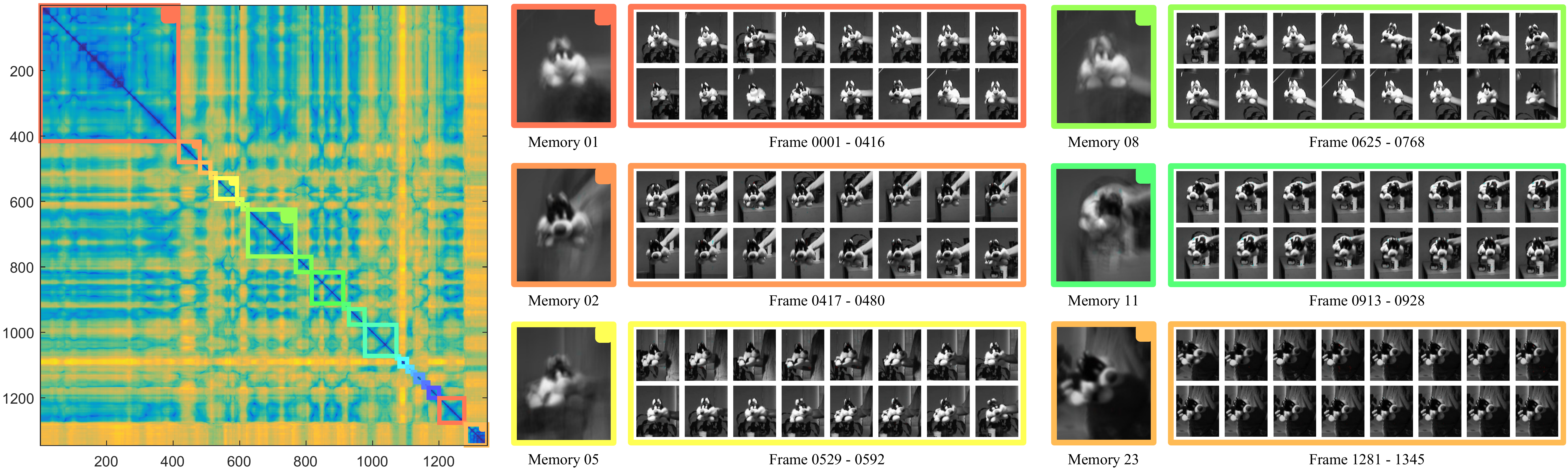}\\
\end{center}
\vspace{-2mm}
\hspace{14.0mm}{\small Distance Matrix and Clustering Result}\hspace{15.0mm}{\small Six temporally constrained clusters with distinct appearances}\\
\vspace{-6mm}
   \caption{
    Left: the distance matrix $\mathbf D$ as described in Alg.~\ref{alg:greedy}.
    Right: Six representative clusters with corresponding colored bounding boxes are shown
    for intuitive understanding.
    The image patches in the big bounding boxes is an average appearance of a certain cluster (memory),
    while the small patches are samples chosen evenly on the temporal domain from each cluster.
   } \label{fig:f2}\vspace{-5mm}
\end{figure*}

\vspace{-1mm}
\section{Proposed Method} \label{sec:sec3}
Aside from the convolution based visual trackers mentioned above, many other trackers~\cite{jepson2003robust,nummiaro2003adaptive,ross-ijcv08,BabenkoYB11} also update their models $\hat {\mathbf Q}$ in similar form as ${\hat{{\mathbf Q}}^p} = (1 - \gamma ){\hat{\mathbf Q}^{p - 1}} + \gamma { {\mathbf Q}^p}$ with a learning-rate parameter $\gamma \in (0, 1]$ and suffers from the drifting problem.

We observe that smooth movements usually offer consistent appearance cues, which can be modeled as reliable memories to recover the tracker from drifting issues caused by drastic appearance change (illustrated in Fig.~\ref{fig:idealpath}). In this section, we introduce our method that explores, preserves and makes use of the reliable memories for long-time robust visual tracking. First, we introduce our novel framework that is capable of handling vast number of samples while still ensures fast detection speed. Then, we elaborate the details of intelligently arranging past samples into distinct and reliable clusters that grant our tracker resistance to drift error.

\vspace{-1mm}
\subsection{The Circulant Tracker over Vast Samples}

Given new positive sample $\boldsymbol x^p$ at frame $p$, we aim to build an adaptive model $\{\hat {\boldsymbol x}^p, \hat {\mathbf A}^p\}$ for fast detection in the coming frame $p+1$ with sample image $\boldsymbol z$ by
\vspace{-1.5mm}
\begin{equation}\label{eq:yp1}\vspace{-1.5mm}
\boldsymbol y^{p+1} = {\mathscr F^{ - 1}}({{\hat {\mathbf A}}^p} \odot \mathscr F({{\tilde {\boldsymbol k}}^{p}})),
\end{equation}
where $\boldsymbol y^{p+1}$ is the response map which shows the estimated translation of the target position, vector ${\tilde {\boldsymbol k}}^{p} \in \mathbb R^L$, with its $l$-th entry $\tilde {\boldsymbol k}_l^p := \kappa (\boldsymbol z,\hat {\boldsymbol x}_l^p)$.
As we advocated, this model $\{\hat {\boldsymbol x}^p, \hat {\mathbf A}^p\}$ should be built upon vast samples  for robustness and adaptiveness.
Thus,  $\hat {\boldsymbol x}^p$ should have the form:\vspace{-2.5mm}
\begin{equation}\label{eq:hatxp}\vspace{-0.5mm}
  {{\hat {\boldsymbol x}}^p} = (1 - \gamma )\sum\limits_{j = 1}^{p - 1} {{\beta ^j}{\boldsymbol x^j} + \gamma {\boldsymbol x^p}} ,\quad \gamma  \in (0,1],\;\sum\limits_{j = 1}^{p - 1} {{\beta ^j}}  = 1.
\end{equation}
As shown, the adaptive learned appearance $\hat {\boldsymbol x}^p$ is a combination of past $p$ samples with concentration on $\boldsymbol x^p$ of a certain proportion $\gamma$.
Coefficients $\{\beta ^j\}^{p-1}_{j=1}$ represent the correlation between the current estimated appearance $\hat {\boldsymbol x^p}$ and the past appearances $\{\boldsymbol x^j\}^{p-1}_{j=1}$.
A proper choice of $\{\beta ^j\}^{p-1}_{j=1}$ should make the model: 1) adaptive to new appearance changes, and 2) consistent with past appearances to avoid risk of drifting.
In this paper, we argue that the set $\{\beta ^j\}^{p-1}_{j=1}$ with preference to
previous reliable memories can provide our tracker with considerable robustness to resist drift error.
We discuss how to find these reliable memories in Sec.~\ref{sec:semi}, and their connections with
$\{\beta ^j\}^{p-1}_{j=1}$ are introduced in Sec.~\ref{sec:workflow}.

Now, we focus on finding a set of classifier coefficients $\boldsymbol \alpha$ that fit both the learned appearance $\hat {\boldsymbol x}^p$ for consistency and the current appearance $\boldsymbol x^p$ for adaptiveness. Based on Eq.~\ref{eq:rls} and Eq.~\ref{eq:simplest_rls}, we derive the following cost function to minimize:\vspace{-2.5mm}
\begin{equation}\label{eq:rlsp}\vspace{-3.5mm}
\begin{split}
F^p(\boldsymbol \alpha) =& (1-\gamma)\left [(\boldsymbol y-\hat {\mathbf K}^p \boldsymbol\alpha)^{\rm T}(\boldsymbol y-\hat {\mathbf K}^p \boldsymbol \alpha)+ \lambda \boldsymbol \alpha^{\rm T}\hat {\mathbf K}^p \boldsymbol \alpha \right ]\\
&+\gamma \left [ (\boldsymbol y- \mathbf K^p {\boldsymbol \alpha})^{\rm T}(\boldsymbol y- \mathbf K^p \boldsymbol\alpha)+ \lambda \boldsymbol\alpha^{\rm T} \mathbf K^p \boldsymbol\alpha  \right],
\end{split}
\end{equation}
where the kernel matrix $\hat {\mathbf K}^p = C(\hat {\boldsymbol k}^p)$, and vector entry $\hat { k}^p_l = \kappa (\hat {\boldsymbol x}^p, \hat {\boldsymbol x}^p_l )$ (similar for $\mathbf K^p$ and $\boldsymbol k^p$).
$\gamma$ is a balance factor between the memory consistency and model adaptiveness.
%
By setting the derivative $\nabla F^p_{\boldsymbol\alpha} = 0$, the accurate solution $\hat {\boldsymbol\alpha}^p$ satisfies a complicated condition as follows:\vspace{-2.5mm}
\begin{equation}\label{eq:deri2}\vspace{-2.0mm}
\begin{split}
  &\left [ (1 - \gamma) \hat {\mathbf K}^p ( \hat {\mathbf K}^p + \lambda \mathbf I) + \gamma \mathbf K^p (\mathbf K^p + \lambda \mathbf I) \right ] \hat {\boldsymbol\alpha} ^p\\
& = \left [ (1-\gamma)\hat {\mathbf K}^p + \gamma {\mathbf K}^p) \right ]\boldsymbol y.
\end{split}
\end{equation}
We observe that the adaptively learned appearance $\hat {\boldsymbol x}^p$ should be very close to the current one
$\boldsymbol x^p$, since it is a linear combination of close appearances in the past $\{{\boldsymbol x^j}\}_{j=1}^{p-1}$ 
and the current appearance $\boldsymbol x^p$, as shown in Eq.~\ref{eq:hatxp}.
Notice both kernel matrix $\mathbf K^p$ and $\hat {\mathbf K}^p$ (and their linear combination with $\lambda \mathbf I$) is positive semidefinite.
By relaxing Eq.~\ref{eq:deri2} with $\hat {\mathbf K}^p \simeq (1-\gamma)\hat {\mathbf K}^p + \gamma \mathbf K^p \simeq \mathbf K^p$,
we obtain an approximate minimizer $\hat {\boldsymbol \alpha}^p$ in a very simple form:\vspace{-2.6mm} 
\begin{equation}\label{eq:simplealpha}\vspace{-2.5mm}
\begin{split}
  \hat {\boldsymbol \alpha}^p &\simeq \left [ (1-\gamma)\hat {\mathbf K}^p + \gamma \boldsymbol K^p + \lambda \mathbf I) \right ]^{-1} \boldsymbol y\\
&= \left[C( (1 - \gamma)\hat {\boldsymbol k}^p + \gamma \boldsymbol k^p + \lambda \boldsymbol \delta) \right ]^{-1} \boldsymbol y\\
&=\mathscr F^{-1}(\frac{\mathscr F(\boldsymbol y)}{(1- \gamma)\mathscr F (\hat {\boldsymbol k}^p) + \gamma \mathscr F (\boldsymbol k^p) + \lambda \mathbbm 1}).
\end{split}
\end{equation}
$\boldsymbol\delta$ is an $L$-dimensional vector in the form $\boldsymbol\delta = [1, 0, ..., 0]^\mathrm{T}$, with property that
$C(\boldsymbol\delta) = \mathbf I$ and $\mathscr F (\boldsymbol\delta) = \mathbbm 1$ ($\mathbbm 1$ is an L-dimension vector of ones).
Note that in the bracket of $\mathscr F^{-1}(\cdot)$, division is performed element-wise.

As long as we find a proper set of coefficients $\{\beta ^j\}^{p-1}_{j=1}$, we can
build up our detection model $\{\hat {\boldsymbol x}^p, \hat {\mathbf A}^p\}$ by Eq.~\ref{eq:hatxp} and Eq.~\ref{eq:simplealpha}. In the next frame $p+1$, fast detection can be performed by  Eq.~\ref{eq:yp1} with this learned model.
%
\begin{algorithm}
\caption{Temporal Constrained Clustering Algorithm }\label{alg:greedy}
\begin{algorithmic}
\STATE \textbf{Input:} Integral image $\mathbf J$ of Distance Matrix $\mathbf D \in \mathbb R^{p \times p}$, s.t.
$\mathbf D_{ij} = ||\phi (\boldsymbol x^i) - \phi (\boldsymbol x^j)||^2, \; \forall i, j = 1,...,p$;\\
$\mathbf M = \{\boldsymbol m_i\}^p_{i = 1}, \; \boldsymbol m_i = \mathbf P^i \boldsymbol\delta, \; \forall i = 1,...,p$;\\
$\mathbf P^i$ is a shifting matrix and $\boldsymbol \delta = [1, 0,..., 0]^{\rm T}$;\\
Stoping factor $\rho$, and $N = |\mathbf M| + 1$.
\STATE \textbf{Output:} $\hat {\mathbf M} = \{\boldsymbol m_i\}^H_{i = 1}$.
\WHILE{$(|\mathbf M|<N)$}
\STATE $N = |\mathbf M|$;
\FOR{$h = 1 : 2 : |\mathbf M| \text{ do}$}
\STATE Evaluate $\tau(\boldsymbol s^h, \boldsymbol s^{h+1}) = \mathcal C(\boldsymbol s^h \bigcap {\boldsymbol s}_{h+1}) - (\mathcal C(\boldsymbol s^h)+\mathcal C(\boldsymbol s^{h+1}))$ using $\mathbf J$.
\IF{$\tau(\mathcal C(\boldsymbol s^h, \boldsymbol s^{h+1})) \le \rho (\mathcal C(\boldsymbol s^h) + \mathcal C(\boldsymbol s^{h+1}))$}
\STATE $\boldsymbol m_h = \boldsymbol m_h + \boldsymbol m_{h+1}$, remove $\boldsymbol m_{h+1}$ from $\mathbf M$;
\ENDIF
\ENDFOR
\ENDWHILE
\STATE $\hat {\mathbf M} = \mathbf M$.
\end{algorithmic}
\end{algorithm}

\subsection{Clustering with Temporal Constraints}
\label{sec:semi}

%

In this subsection, we introduce our temporally constrained clustering, which learns distinct and reliable memories from the incoming samples in a very fast manner. 
Together with the ranked memories (Sec.~\ref{sec:workflow}), our tracker is robust to inaccurate tracking result, and is able to recover from drift error.

Suppose a set of positive samples are given at frame $p$: $\mathbf S = \{\boldsymbol x^i \}_{i = 1}^p$, and
we would like to divide them into $H$ subsets $\{\boldsymbol s^h\}_{h = 1}^H$
with indexing vector set $\mathbf M = \{\boldsymbol m^1,...,\boldsymbol m^H \} \in \{0, 1\}^p$, such that
 $\boldsymbol s^h:=\{\boldsymbol x^i: m^h_i=1,\forall i = 1,...,p\}$.
Our objective are as follows: 1) Samples in each subset $\boldsymbol s^h$ are highly correlated; 2) Samples from different subsets have relatively
large appearance difference, so a linear combination of them is vague or even ambiguous to describe
the tracked target (\emph{e.g.}, samples from different viewpoints of the target).
Thus, it can be modeled as a very general clustering problem:\vspace{-4.5mm}
\begin{equation}\label{eq:cluster}\vspace{-3mm}
\begin{split}
  \min_{\mathbf M}  \sum_h&\mathcal C(\boldsymbol s^h) + \eta r(|\mathbf M|),\\
s.t.\; \langle \boldsymbol m^i, \boldsymbol m^j \rangle &= 0,\;\forall \boldsymbol m^i, \boldsymbol m^j \in \mathbf M,\; i\ne j;\\
\sum_h {\boldsymbol m^h} &= \mathbbm 1_{p \times 1}.
\end{split}
\end{equation}
Function $\mathcal C(\boldsymbol s^h)$ measures the average sample distance in feature space $\phi ( \cdot )$
within subset $\boldsymbol s^h$, in the form: $\mathcal C(\boldsymbol s^h)=({ {\mathbbm 1_{p \times 1}} ^{\rm T} \times {\boldsymbol m^h} })^{-1} \sum_{\forall \boldsymbol x^i,\boldsymbol x^j \in \boldsymbol s^h, i < j}||\phi(\boldsymbol x^i)-\phi (\boldsymbol x^j)||^2$.
Regularizer $r(|\mathbf M|)$ is a function based on the number of subsets $|\mathbf M|$, and $\eta$ is a balance factor.
This is a discrete optimization problem and known as NP-hard.
By fixing the number of subsets $|\mathbf M|$ to a certain constant $k$, $k$-means clustering can converge to a local optimal.

However, during the process of visual tracking, we do not know the sufficient number of clusters.
While too many clusters cause problem of over-fitting, too few clusters may lead to ambiguity.
More importantly, as long as we allow random combinations of samples during clustering, any cluster has a risk of taking in contaminated samples with drift error, even wrongly labeled samples, which in turn will degrade the performance of models built upon them.

One important observation is that, target appearances closed to each other in the temporal domain may form a very distinguished and consistent pattern, \emph{i.e.}, reliable memories. \emph{E.g,}, if a well-tracked target moves around without big rotation or large change in angle for a period of time, its edge-based feature would have much higher similarity compared with feature under different angles. 
In order to discover these memories, we add temporal constraints on Eq.~\ref{eq:cluster}:\vspace{-1mm}
\begin{equation}\label{eq:constraint}\vspace{-0.mm}
\begin{split}
\boldsymbol m^h &\in \mathbf T,\; \forall h=1,...,H;\\
\mathbf T :=\{\boldsymbol t\in\{0, 1\}^p &: \text{all $t_i = 1$ are concatenated.}\}.
\end{split}
\end{equation}
Then Eq.~\ref{eq:cluster} with Eq.~\ref{eq:constraint} becomes segmenting 
$\mathbf S$ into subsets $\{\boldsymbol s^h\}_{h=1}^H$, that each subset only contains timely continuous samples
$\boldsymbol s^h = \{\boldsymbol x^i\}^v_{i = u}$ ($u, v$ are certain frame numbers).

Still, the constraint of this new problem is discrete and the global optimal can hardly be reached.
We carefully designed a greedy algorithm, as shown in Alg.~\ref{alg:greedy}, which starts from a trivial status of $p$ subsets.
It tries to reduce the regularizer $r(|\mathbf M|)$ in the object function
of Eq.~\ref{eq:cluster} by combining temporally adjacent subsets $\boldsymbol s^h$ and $\boldsymbol s^{h+1}$,
while penalizing the increase of the average sample distance
$\tau(\boldsymbol s^h, \boldsymbol s^{h+1}) := \mathcal C(\boldsymbol s^h \bigcap {\boldsymbol s}_{h+1}) - (\mathcal C(\boldsymbol s^h)+\mathcal C(\boldsymbol s^{h+1}))$.

With an intelligent use of Integral Image~\cite{viola2001rapid}, the evaluation operation in each combining step in Alg.~\ref{alg:greedy} only
takes $O(1)$ running time with integral image $J$, and each iteration takes linear $O(p)$ operations. 
The whole algorithm processes in a bottom-up binary tree structure, and runs at
$O(p\log p)$ in the worst case, and runs less than $30$ms on a desktop for over $1000$ samples.
Designed experiments will show that the proposed algorithm
is very competent in finding distinguished appearance clusters (reliable memories) for our
tracker to learn.

\subsection{The Workflow of Our Tracking Framework}\label{sec:workflow}

Two feature pools are employed in our framework, one for coming positive samples across frames, and the second (
denoted by $\mathbf U$)
for the learned memories.
Every memory $\boldsymbol u \in \mathbf U$ contains a certain number of samples $\{\boldsymbol x^{\boldsymbol u}_j\}_{j = 1}^{N^{\boldsymbol u}}$
and a confidence $c_{\boldsymbol u}$:\vspace{-1mm}
\begin{equation}\label{eq:mconf}\vspace{-1mm}
  c_{\boldsymbol u} = e^{-(\sigma_1B^{\boldsymbol u} - \sigma_2N^{\boldsymbol u})},
\end{equation}
where $N^{\boldsymbol u}$ is the number of samples in memory $\boldsymbol u$ and $B^{\boldsymbol u}$ is the beginning  frame number of memory $\boldsymbol u$.
This memory confidence
is consistent with our hypothesis that earlier memories with more samples are more stable and less likely to be affected by accumulated drift error.
For each frame, we first detect the object using Eq.~\ref{eq:yp1} to estimate
the translation of the target, and then utilize this new sample to update our appearance model
$\{\hat {\boldsymbol x}^p, \hat {\mathbf A}^p\}$ by Eq.~\ref{eq:hatxp} and Eq.~\ref{eq:simplealpha}.

The correlation coefficient $\{\beta^j\}_{j = 1}^{p-1}$ is then calculated by:\vspace{-2mm}
\begin{equation}\label{eq:beta}\vspace{-2mm}
 \beta^j = \Theta^{-1}\exp\{-\sum_{j\in \hat{\boldsymbol u}}||\phi (\boldsymbol x^p) - \phi (\boldsymbol x^j)||^2\},
\end{equation}
where scalar $\Theta$ is a normalization term that assures $\sum_{j = 1}^{p-1}{\beta^j} = 1$, and $\hat{\boldsymbol u}$
is the  most similar memory to the current learned appearance $\hat {\boldsymbol x}^p$ in feature space $\phi(\cdot)$.

To update memories, we use Alg.~\ref{alg:greedy} to cluster positive samples in the 
first feature pool into `memories', and import all except the last one  into $\mathbf U$. Note when $|\mathbf U|$
reaches its threshold, memories with the lowest confidence would be abandoned immediately.


\section{Experiments}

Our framework is implemented in Matlab with running speed ranges
from $12$fps to $20$fps, on a desktop with an Intel Xeon(R) 3.5GHz CPU, a Tesla K40c video card and 32GB RAM.
The adaptiveness ratio $\gamma$ is empirically set as $0.15$ through all experiments.
Stoping factor $\rho$ is decided adaptively as $1.2$ times the average covariance of the samples
at the first $40$ frames on each video. HOG~\cite{hog} is chosen as the feature $\phi(\cdot)$.
The maximum number of memories $|\mathbf U|$ is set as $10$ and $max(N^{\boldsymbol u})$ is set to $100$.
%
\begin{figure*}[htbp]
\begin{center}
   \includegraphics[width=0.85\linewidth]{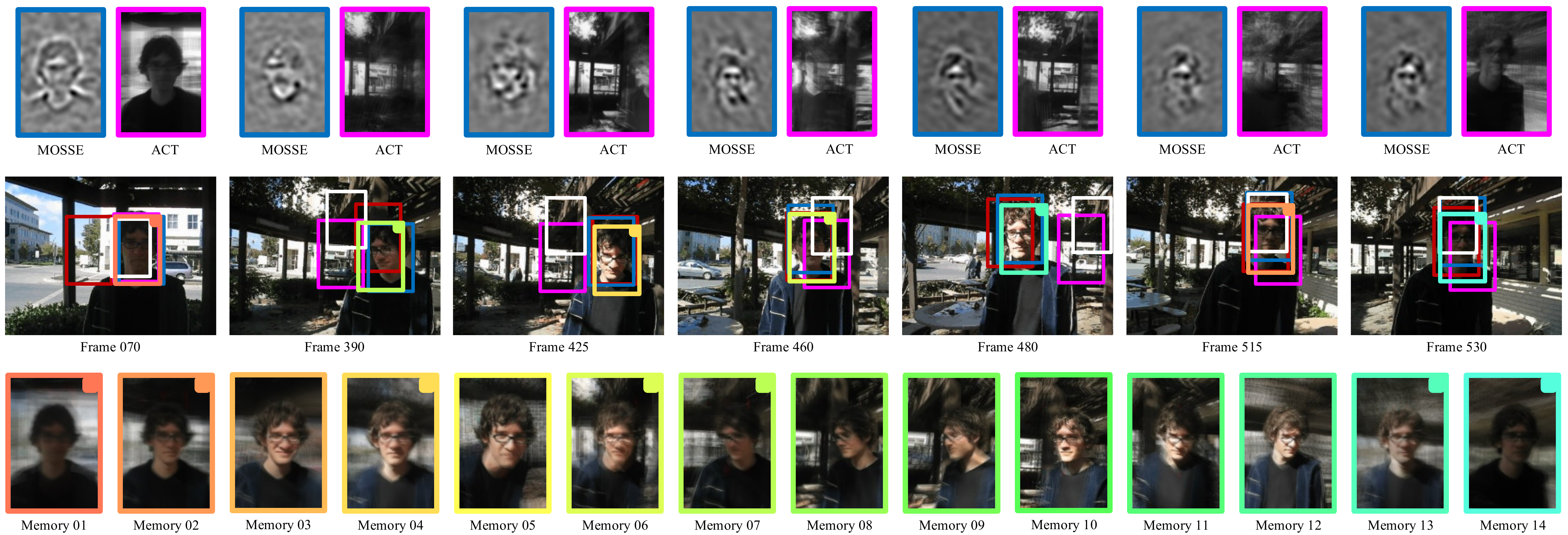}\\
\end{center}
\vspace{-1mm}
\vspace{-4mm}
   \caption{Comparison on the $Trellis$ video. Blue, magenta, red and white
   represent results from MOSSE, ACT, Struck and VTD. Our results are represented as bold
   colored boxes with a dot on the top-right corner, and each color means the active memory at that frame, shown in the bottom row.
   Learned appearances from MOSSE and ACT at each frame are also shown in the top row.
   As illustrated, the target experiences out-of-plane rotation in frame $390$ and brings about drift error (memory $07$). When he turns head back to the front (frame $425$, $460$), our method uses reliable memories $04$ and $06$ respectively to recover from drifting. Note that these two memories were built before drift error accumulates at out-of-plane rotation period.
   } \label{fig:figure3}\vspace{-1mm}
\label{fig:f3}
\end{figure*}
\subsection{Evaluation of Temporally Constrained Clustering}

In order to validate our assumption that temporally constrained clustering on sequentially tracked samples
forms reliable and recognizable patterns, we perform Alg.~\ref{alg:greedy} on the off-line positive samples
based on our tracking results. Note that our algorithm gives exactly the same result in the online/offline manner,
since previously clustered samples have no effect on clustering the unfixed sample afterwards. 
Due to space limitation, here we present illustrative results from sequences $Sylvester$ and $Trellis$, in Fig.~\ref{fig:f2} and
Fig.~\ref{fig:f3}.

Fig.~\ref{fig:f2} shows our detailed results on sequence $Sylvester$, in which
the target experiences illumination variation, in-plane and out-of-plane rotation
through a long term of $1345$ frames.
The left part shows the distance matrix $\mathbf D$ as described in Alg.~\ref{alg:greedy},
that $\mathbf D_{ij} = ||\phi (\boldsymbol x^i) - \phi (\boldsymbol x^j)||^2, \; \forall i, j = 1,...,p$.
Pixel $\mathbf D_{ij}$ with dark blue (light yellow) implies small (large) distance between sample $\boldsymbol x^i$ and $\boldsymbol x^j$
in feature space $\phi (\cdot)$.
Different colored diagonal bounding boxes represent different temporally constrained clusters.
The right part shows six representative clusters, corresponding to colored bounding boxes on the matrix.
Memory $\#1$ and memory $\#8$ are two large clusters containing large amount of samples with
high correlated appearance (blue color).
Memory $\#11$ represents a cluster with only $16$ samples. Its late emergence and limited number of samples
result in a very low confidence $c_{\boldsymbol u}$ and thus it is not likely to replace any existing reliable memories.


Fig.~\ref{fig:f3} shows mid-term results on the sequence $Trellis$, compared with two very related trackers
MOSSE and ACT, with their learned appearances at these frames.
Several other state-of-the-art methods are also shown for comparison.
Among them, MOSSE and ACT only keep one memory, which is generated from former appearances
and gradually adapts to the current one.
Though they are very robust to illumination changes, drift error still accumulates across frames.
They can hardly recover from drifts due to the conciseness of their model.
Our method, with learned reliable memories, is very robust to appearance changes,
and can recover from drifts when it observes appearances familiar to its memories.

\vspace{-1mm}
\subsection{Boosting by Deep CNN}
Our tracker's inherent requirement to efficiently search familiar patterns (memories) at a global scale of the frame overlaps with object detection task~\cite{girshick2015ICCV,he2015spatial}. Recently, with the fast development of convolutional neural networks (CNN)~\cite{Krizhevsky2012NIPS,zeiler2014eccv}, Faster-RCNN~\cite{renNIPS15fasterrcnn} achieves $\ge$ $5$ fps detection speed by using shared convolutional layers for both the object proposal and detection.
To equip our tracker with a global vision for its reliable memories, we fine-tune the FC-layers of a Faster-RCNN detector (ZF-Net) once we have learned sufficient memories in a video, which helps our tracker resolve local minimum issues caused by limited effective detection range. 
Though only supplied with coarse detections with a risk of false alarms, our tracker can start from a local region close to the target and then ensure accurate and smooth tracking results.
Note that we only tune the CNN once, with around 150 seconds running time on one Tesla K40c for 3,000 iterations. When the tracking task is long, \emph{e.g.}, more than 3,000 frames, the average fps is larger than $15$, which is certainly worthy for significant improvement in robustness.
In the following stage, we perform CNN detection every $5$ frames, each taking less than $0.1$s.
\begin{figure*}[htbp]
\begin{center}
   \includegraphics[width=0.185\linewidth]{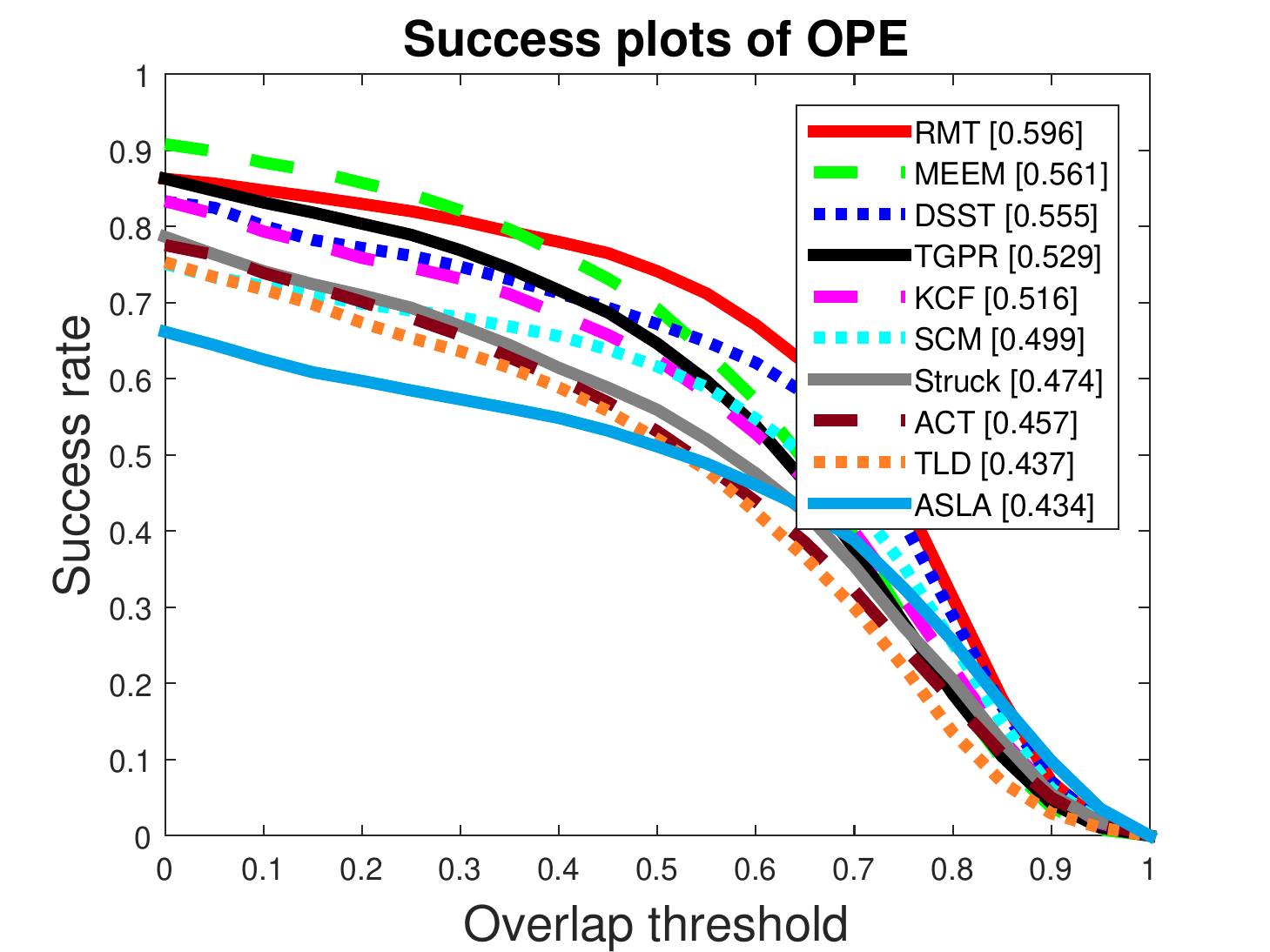}
   \includegraphics[width=0.185\linewidth]{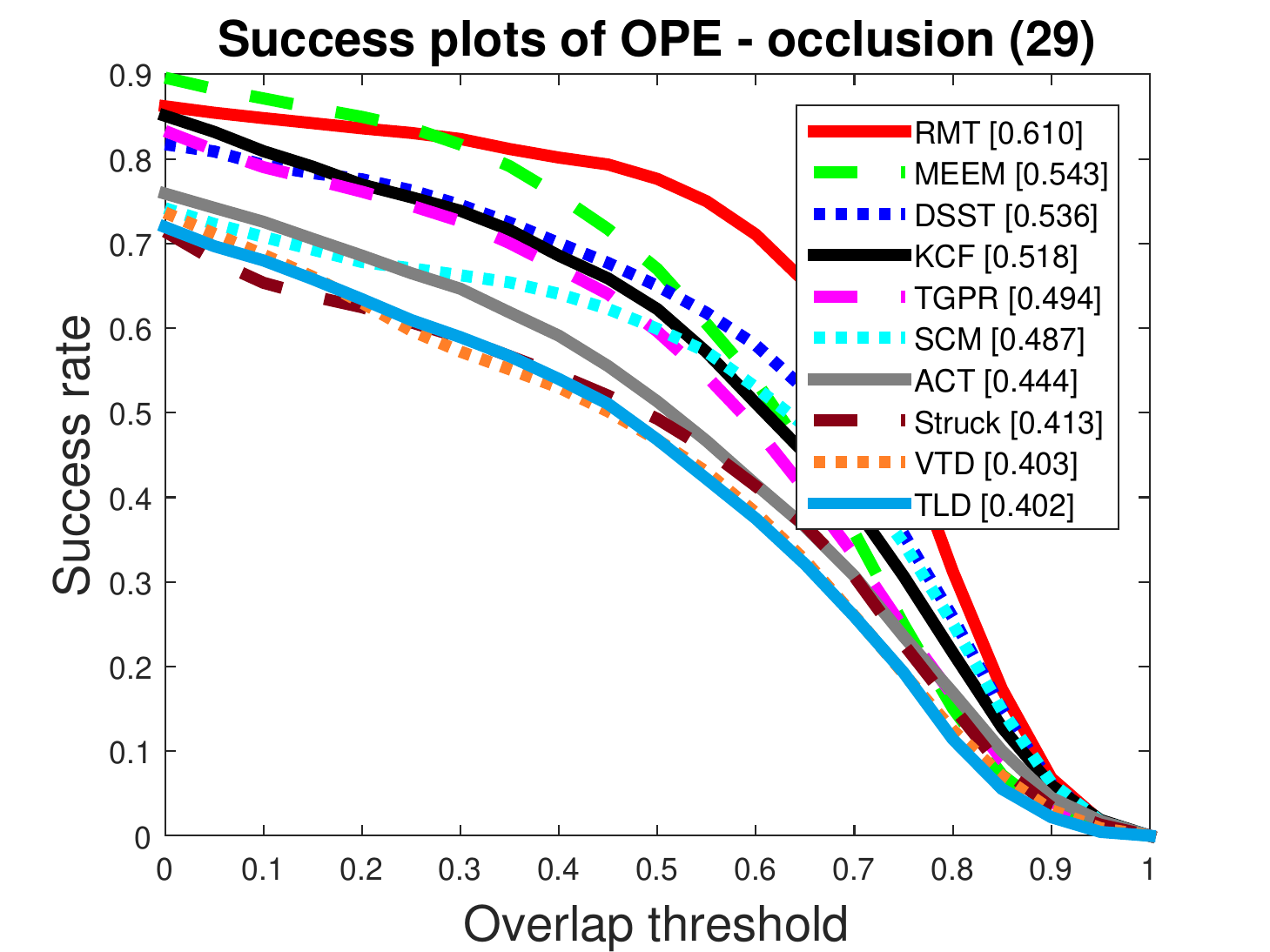}
   \includegraphics[width=0.185\linewidth]{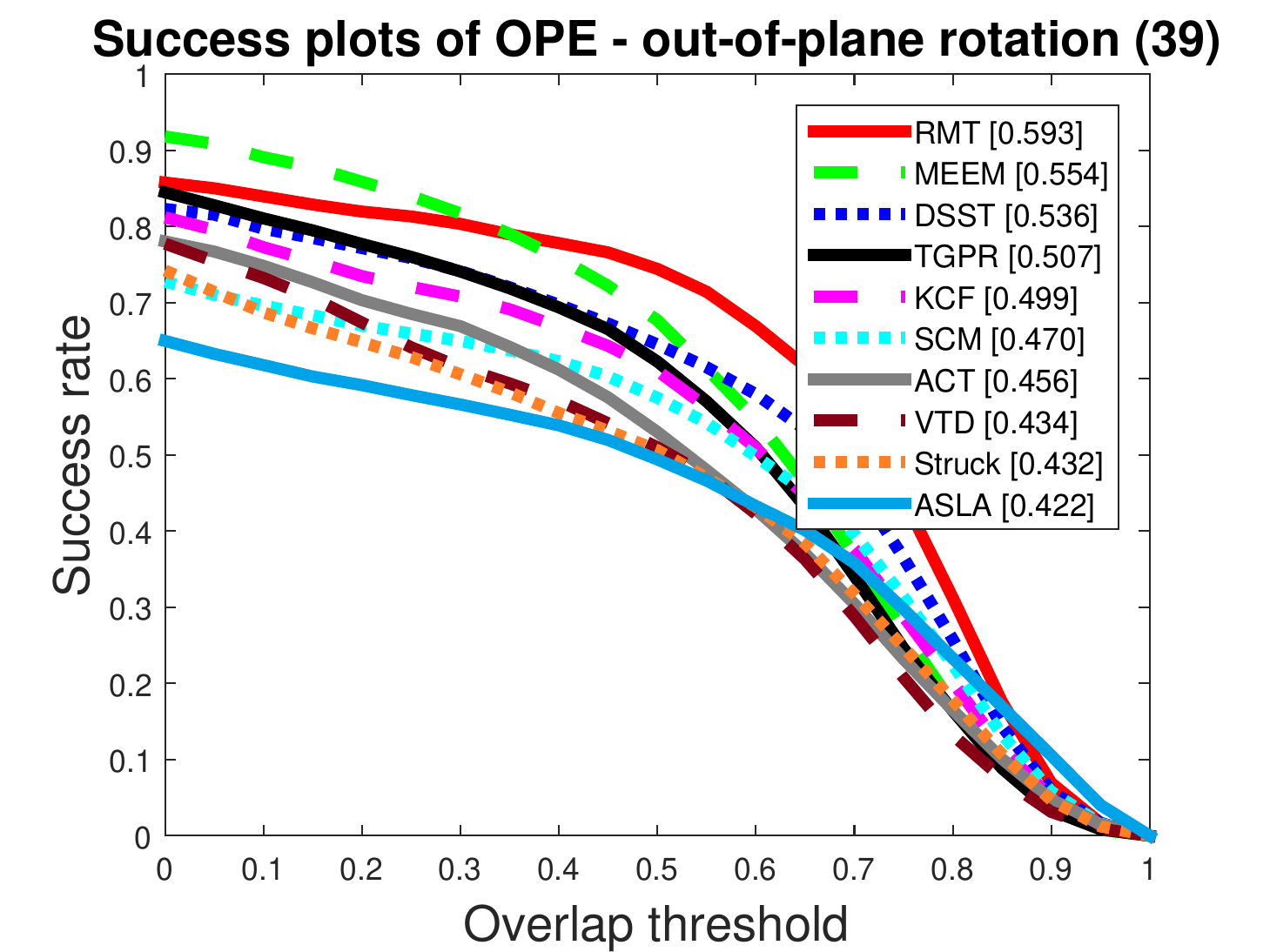}
   \includegraphics[width=0.185\linewidth]{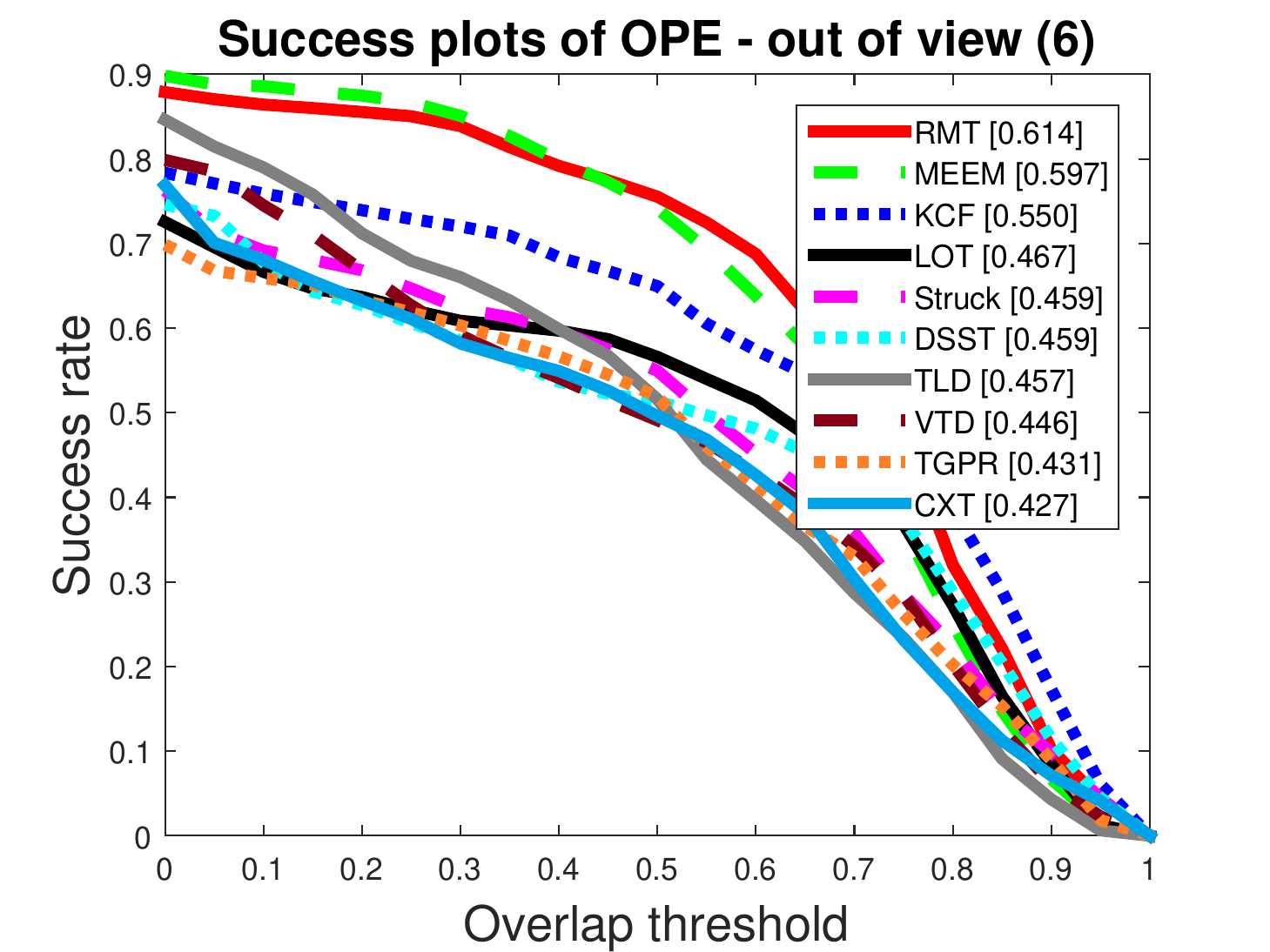}
   \includegraphics[width=0.185\linewidth]{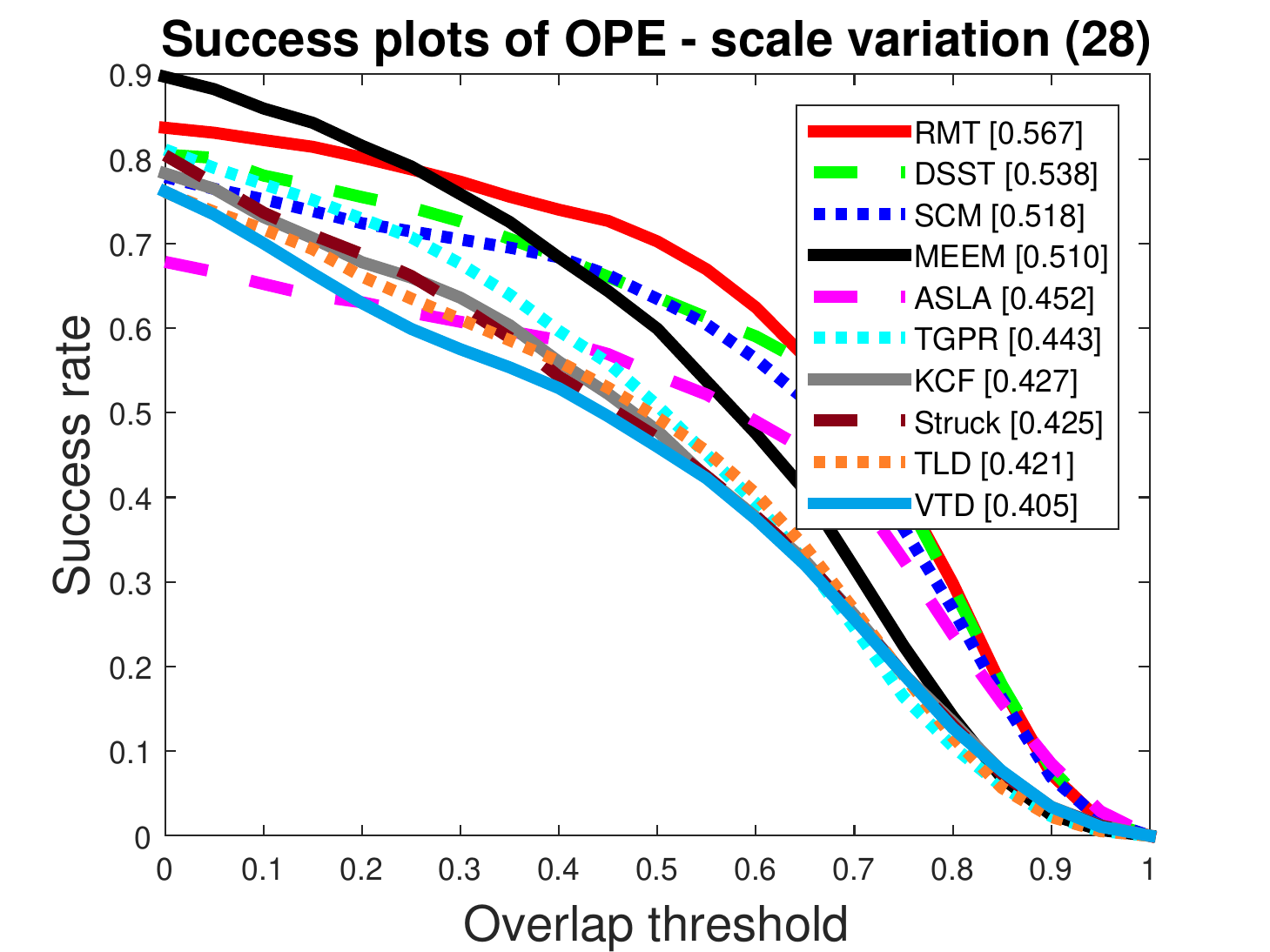}
\end{center}
\vspace{-5mm}
   \caption{Tracking result comparison on $50$ sequences from the OTB-2013 dataset.
    Our tracker is represented by RMT and achieved the top performance on success plots evaluation standard.
   MEEM, TGPR, DSST and KCF also have close performance to our tracker. Only top-$10$
   out of $14$ tracker results are shown for clearness.  
   } \label{fig:ope2}\vspace{-4mm}
\end{figure*}

\vspace{-1mm}
\subsection{Quantitative Evaluation}


We first evaluate our method on $50$ challenging sequences from
OTB-2013~\cite{WuLimYang13} against $14$ state-of-the-art methods:
ACT~\cite{nummiaro2003adaptive},
ASLA~\cite{jia2012visual},
CSK~\cite{Henriques-eccv12},
CXT~\cite{dinh2011context},
DSST~\cite{danelljan2014accurate},
KCF~\cite{henriques2015tpami},
LOT~\cite{oron2012locally},
MEEM~\cite{zhang2014meem},
MOSSE~\cite{Bolme-cvpr10},
SCM~\cite{zhong2012robust},
Struct~\cite{hare2011struck},
TGPR~\cite{gao2014transfer},
TLD~\cite{kalal2012tracking}
VTD~\cite{kwon2010visual}.
We employed the released code from the public resource (\emph{e.g.}, OTB-2013) or the released version by
the authors, and all parameters are fixed for each trackers during testing.
Fig.~\ref{fig:ope2} shows the success plots on the whole dataset with the one pass evaluation (OPE) standard.
Our tracker, represented as RMT (Reliable Memory Tracker), obtains the best performance, while MEEM, TGPR, KCF and DSST also provide
competitive results.
TGPR's idea of building one tracker on auxiliary (very early) samples and MEEM's idea of
using tracker's snapshot can be interpreted as making use of early formed reliable memory patterns,
which is very relevant to our method.
DSST designs a very concise pyramid representation for object scale estimation, and employs robust dense HOG
feature to achieve high accuracy on estimating target motion.
Our tracker outperforms the others on most challenging scenarios, \emph(e.g.), occlusion, out-of-plane rotation, out of view, fast motion,
as illustrated by Fig.~\ref{fig:ope2}. 
The main reason is that our tracker possesses amount of very reliable memories and a global vision that help
it regain focus on the target after drastic appearance changes.\vspace{-3mm}
%
\begin{table}[h!]\scriptsize
  \begin{center}
    \begin{tabular}{l|@{\hspace{1em}}c@{\hspace{1em}}c@{\hspace{1em}}c@{\hspace{1em}}c@{\hspace{1em}}c@{\hspace{1em}}c@{\hspace{1em}}c@{\hspace{1em}}c}
      \toprule
      Sequence      &Fr. No.   & MOSSE &  KCF   &  ACT    & DSST      & TLD       & MEEM &  RMT\\
      \midrule
      \midrule
      Motocross     & 2,035   & 295.9  & 181.5  & 182.5   &    67.5   &  44.7     &  \emph{33.4}  &  \textbf{21.5}\\
      Volkswagon    & 4,000   &  60.6  & 114.1  &  41.3   &    122.7  &  \emph{15.9}     &  51.1  &  \textbf{12.3}\\
      Carchase      & 4,000   &  125.0 & 129.4  &  98.0   &    132.6  &  \emph{34.4}     &  38.1  &  \textbf{34.1}\\
      Panda         & 3,000   &  64.8  & 83.3   &  64.5   &    71.4   &  \emph{27.1}     &  97.9  &  \textbf{23.9}\\
      \midrule
      \textbf{Overall}    & 13,035    &  118.5 & 122.3    & 86.1    &  105.3  &  \emph{28.7} &  55.1   &  \textbf{23.1}   \\
      \bottomrule
    \end{tabular}
  \end{center}\vspace{-4mm}
  \caption{Tracking result comparison based on average errors of center location in pixels (the smaller the better) on
                        four long-term videos over $13,000$ frames. Average performances are weighted by the frame number for fairness.}
  \label{tab:table1}
\end{table}\vspace{-2mm}

In order to explore the robustness of our tracker, and validate its resistance to drift error on
long-term challenging tasks, we run our tracker on four long sequences from~\cite{kalal2012tracking}, over 13,000 frames in total.
We have also evaluated the convolution filter based methods that are highly related to our method:
MOSSE~\cite{Bolme-cvpr10}, KCF~\cite{henriques2015tpami}, ACT~\cite{nummiaro2003adaptive} and DSST~\cite{danelljan2014accurate},
together with MEEM~\cite{zhang2014meem} and a detector-based method TLD~\cite{kalal2012tracking} (shown in Tab.~\ref{tab:table1}).
In order to make fair comparison, we have re-labeled the initial frame to ensure that no tracker lose focus in the beginning.
While MOSSE often loses track at very early frames, KCF, ACT and DSST are able to track the target stably for hundreds
of frames, but usually cannot maintain their focus after 600 frames. MEEM performs favorably on video $Motocross$ for over $1700$
frames with its impressive robustness, but it is unadaptable to scale changes and still leads to inaccurate results.
Our tracker and TLD performs over the other five trackers on all videos since both of them have a global vision to search for the
target.
However, based on an online random forest model, TLD takes in false positive samples slowly, which finally leads to false detections and
inaccurate tracking results. Contrarily, guided by the CNN detector trained with our reliable memories, our tracker is only affected by very
limited number of false detections. It robustly tracks the target across all frames, and gives accurate target location and target scale until the last frame for all four videos.
A video clip with more detailed illustration and qualitative comparison can be found \href{https://youtu.be/wtZAGzFDjnM}{here}.
\vspace{-1mm}

\vspace{-1mm}
\section{Conclusion}

In this paper, we propose a novel tracking framework, which explores temporally correlated appearance clusters across tracked samples, and then preserves reliable memories for robust visual tracking.
A novel clustering method with temporal constraints is carefully designed to help our tracker retrieve good
memories from a vast number of samples for accurate detection, while still ensures its real-time performance.
Experiment shows that our tracker performs favorably against
other state-of-the-art methods, with outstanding ability
to recover from drift error in long-term tracking tasks.

\newpage

\bibliographystyle{named}
\footnotesize{
\bibliography{mmr_track}
}
\end{document}